\documentclass[runningheads]{llncs}
\usepackage{blindtext}

\usepackage[T1]{fontenc}
\usepackage{graphicx}
\usepackage{svg}
\begin{document}
\title{Keypoint-based Diffusion for Robotic Motion Planning on the NICOL Robot}
\titlerunning{Key-Point-based Diffusion for Robotic Motion Planning}
%

\author{Lennart Clasmeier$^\dagger$\inst{1} \and
Jan-Gerrit Habekost$^\dagger$\inst{1}\and
Connor Gäde\inst{1}\and
Philipp Allgeuer\inst{1}\and
Stefan Wermter\inst{1}
\thanks{The authors gratefully acknowledge funding from Horizon
Europe under the MSCA grant agreement No 101072488 (TRAIL).}}
\authorrunning{L. Clasmeier et al.}
%
\institute{Knowledge Technology, Dept. of Informatics, University of Hamburg, Hamburg, Germany
\email{\{jan-gerrit.habekost,connor.gaede,philipp.allgeuer,stefan.wermter\}\\
@uni-hamburg.de}\\
\url{www.knowledge-technology.info}\\
$^\dagger$These authors contributed equally to this work.}

\maketitle              
\begin{abstract}
We propose a novel diffusion-based action model for robotic motion planning. 
Commonly, established numerical planning approaches are used to solve general motion planning problems, but have significant runtime requirements.
By leveraging the power of deep learning, we are able to achieve good results in a much smaller runtime by learning from a dataset generated by these planners.
While our initial model uses point cloud embeddings in the input to predict keypoint-based joint sequences in its output, we observed in our ablation study that it remained challenging to condition the network on the point cloud embeddings.
We identified some biases in our dataset and refined it, which improved the model's performance.
Our model, even without the use of the point cloud encodings, outperforms numerical models by an order of magnitude regarding the runtime, while reaching a success rate of up to 90\% of collision free solutions on the test set. 


\keywords{Motion Planning \and Diffusion Networks \and Humanoid Robots}
\end{abstract}
\section{Introduction}
Robotic motion planning is the process of finding a collision-free plan from a start configuration to a goal configuration.
With the emergence of robotic agents, which are supposed to act autonomously in changing environments, the safety and adaptability of these machines become a core concern.
Numerical motion planning algorithms could help to overcome this task by offering high adaptability to unknown environments, with theoretical guarantees for the optimality and completeness of the plans.
These guarantees, however, come at high computational costs, making near real-time applications impossible and rendering these algorithms unfeasible in interactive scenarios.
Neural motion planners propose a solution by replacing the lengthy planning process with fast neural inference.
Although recent research in the field shows promising results \cite{fishman2022motionpolicynetworks,dalal2024neuralmpgeneralistneural,zang2023graphmp,carvalho2023motion}, motion planning remains an open task, since current approaches are specific to robots and environments, and data generation is very costly.

This work proposes and explores a new diffusion-based approach to neural motion planning on the NICOL platform.
The model approaches the task of motion planning by learning from a synthetic dataset generated by numeric motion planners to create similar plans.
Our initial approach uses point cloud encodings representing the environment at inference time; Ablation experiments reveal that these do not lead to a higher success rate, but slightly decrease the path lengths.
By facilitating spatio-temporal diffusion in the joint space of the robot, we are able to generate 16-step action sequences in a single diffusion run.
By reducing the plans to keypoint representations, 16 steps are enough to generate a plan in one go, which speeds up the planning time significantly.
Further, we apply a batched planning approach, utilizing the parallel inference of the GPU to predict multiple plans for the same task at once, which further stabilized the models' performance and allowed us to reach a 90\% success rate of collision free plans on unseen in-distribution data.

\section{Related Work}

\subsubsection{Neural Motion Planning}
Early neural approaches utilize bio-inspired topological networks with handcrafted neural dynamics to solve low-dimensional dynamic motion planning problems without any learnable parameters \cite{yang2000efficient}.
Ichter et al. \cite{ichter2019learningsamplingdistributionsrobot} introduce a learning sampler that generates candidate configurations for a plan as seed vertices in the data structure of numerical motion planning methods to speed up convergence.
Motion planning networks \cite{qureshi2020motion}, which strongly inspired this work, introduce a behavior-cloning approach by utilizing an MLP-based action predictor and a point cloud encoder to generate trajectories iteratively.
We build upon this approach by replacing the MLP action predictor with a diffusion architecture, which has shown promising capabilities in the field of behavior cloning.
Motion policy networks \cite{fishman2022motionpolicynetworks} are based on a similar behavior cloning approach, but use point cloud-based representations of the robot and the target.
Neural MP \cite{dalal2024neuralmpgeneralistneural} extends the aforementioned approach by utilizing more diverse scenes and an updated architecture.
Khan et al. \cite{khan2020graph} propose graph neural network to predict critical nodes for an RRT-based motion planner.
 The more sophisticated GraphMP \cite{zang2023graphmp} trains two GNNs. One GNN serves as a neural collision checker and the other predicts a graph in the configuration search space. Both networks are trained truth plans and collision data aggregated from standard methods. 
CPP Flow \cite{morgan2023cppflow} utilizes the null space of the IK-Flow inverse kinematics approach to find collision-free joint space trajectories for a given end-effector trajectory.
Zhang et al. show how visuo-motor architectures can be used for motion planning of Cartesian robots in 2D and 3D space \cite{Zang22}, by predicting a probability map of the robot's position for the next image frame.



\subsubsection{Numerical Motion Planning}

Robotic motion planning is a well-known problem that is traditionally approached through sampling-based graph search, as done by probabilistic road maps (PRM) \cite{PRM} and Rapidly-exploring random trees (RRT) \cite{lavalle1998rapidly}. Performance and runtime can be drastically reduced by introducing domain-specific knowledge through heuristics such as $A^*$ \cite{Astar}.
Karaman and Frazzoli \cite{Karaman2011PRM*} show that an adaptive data structure that is extended with new vertices leads to stochastic optimality without increasing the asymptotic runtime.
Informed RRT* \cite{gammell2014informed} limits the search space after an initial solution is found to speed-up convergence.
BIT* \cite{Gammell_2015} starts the sampling process with a limited search space and grows it over the runtime.
AIT* \cite{Strub_2020} introduces an adaptive heuristic to further reduce the search space and runtime.

\subsubsection{Diffusion for Robotic Control}
DALL-E-Bot \cite{10114570} utilizes a diffusion model to create target scene views in a robotic manipulation task.
Diffusion Policy \cite{chi2023diffusionpolicy} introduces diffusion for joint control, using it to generate task-specific robot trajectories conditioned on the robot state and an image encoding of the environment.
Diffusion models have proven themselves in behavior cloning by stabilizing the mapping of a single input to multiple equally valid outputs by utilizing the noise space.
Therefore, we follow this approach in our work as an action generator.
$ \pi 0$ \cite{black2024pi0visionlanguageactionflowmodel} extends a medium-sized Llama architecture by a diffusion head to train a language-conditioned multi-policy agent.  
Octo \cite{octomodelteam2024octoopensourcegeneralistrobot} also uses a diffusion head approach for action generation but introduces a multi-modal Octo transformer to generate an embedding for the diffusion head.
Spisak et al. \cite{spisak2024diffusing} explore the capabilities of diffusion for robotic imitation learning by predicting a first-person view of a robot from a third-person view of the same robot, demonstrating the task.
Carvalho et al. \cite{carvalho2023motion} use a diffusion model as a sampling prior for motion planning, generating plans from noise, without any environmental perception in the input, similar to our ablation study.

\section{Approach} 
We approach the motion planning problem from a behavior-cloning perspective by training a neural network to predict plans from a dataset in a supervised way.
The neural architecture is based on the diffusion policy approach, but instead of using image features from an image encoder, we also initially considered the embeddings from a point cloud encoder.
The diffusion model uses a start and a goal configuration in the robot's joint space and a point cloud embedding of the scene to generate the next 16 steps along a collision-free path to the goal.

\subsection{The NICOL Robot}
As a robotic platform, we use the NICOL (Neuro-Inspired COLlaborator) robot \cite{kerzel2023nicol} in its tabletop setting.
NICOL is a robotic platform designed for the application of machine learning algorithms in human-robot interaction and manipulation scenarios.
NICOL has two 8-DOF Manipulators with anthropomorphic hands to solve challenging manipulation tasks.
The head holds two 4K fisheye cameras; additionally, the robot frame holds multiple depth sensors that scan the workspace from different angles.
The torso of the NICOL robot is fixed in a tabletop setting with the arms mounted 40\,cm above the tabletop.

\subsection{Dataset}
\label{sec:dataset}
Since the plans depend on the robot's layout, we had to create a custom dataset for our scenario.
The dataset consists of 5000 scenes with 20 plans, resulting in 100000 plans.
We use MoveIt to generate ground truth plans in randomly generated scenes with 3 or 4 cuboids of varying sizes.
The positions and rotation of the objects are sampled from the Cartesian workspace bounds of the right arm, and spaced at least 30 cm apart.
The start and the goal poses for the plans are generated via CycleIK \cite{cycleik} with a fixed side-grasp orientation within the workspace bounds, and rejection sampling is applied to ensure the validity of the poses before planning.
For each scene, 20 plans are generated with a planning time of 20 seconds.
We found that shorter planning times led to high failure rates during data generation, due to the numerical planner not finding collision-free plans.
The point clouds are generated synthetically by sampling points from the cuboid planes.

\begin{figure}[t]
\noindent
\begin{minipage}{0.245\linewidth}
\includegraphics[width=\linewidth]{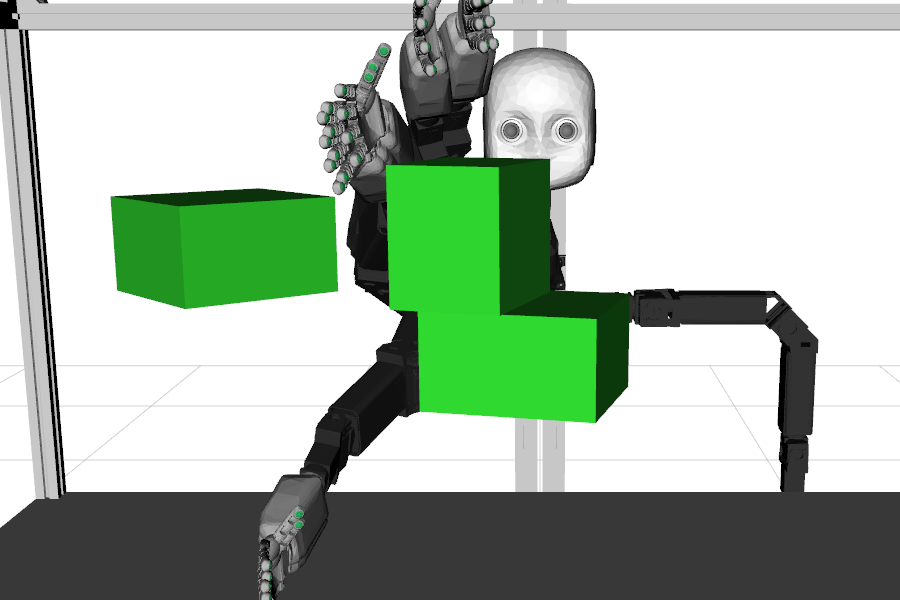}
\centering
\end{minipage}
\begin{minipage}{0.245\linewidth}
\includegraphics[width=\linewidth]{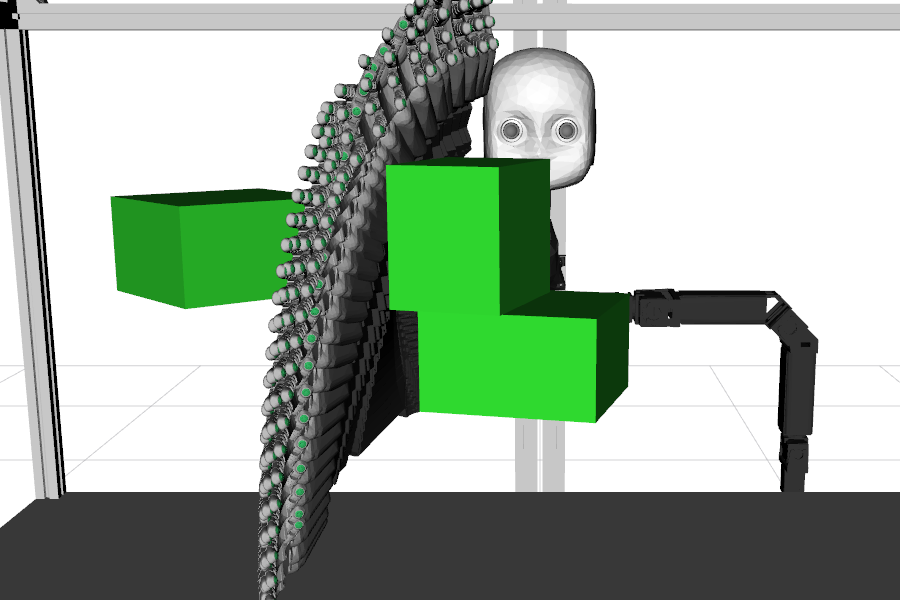}
\centering
\end{minipage}
\begin{minipage}{0.245\linewidth}
\includegraphics[width=\linewidth]{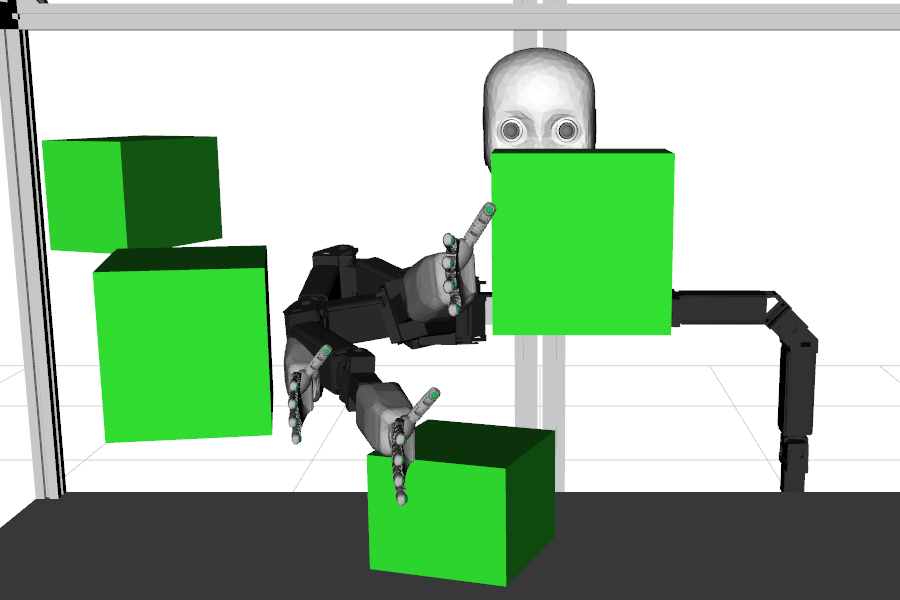}
\centering
\end{minipage}
\begin{minipage}{0.245\linewidth}
\includegraphics[width=\linewidth]{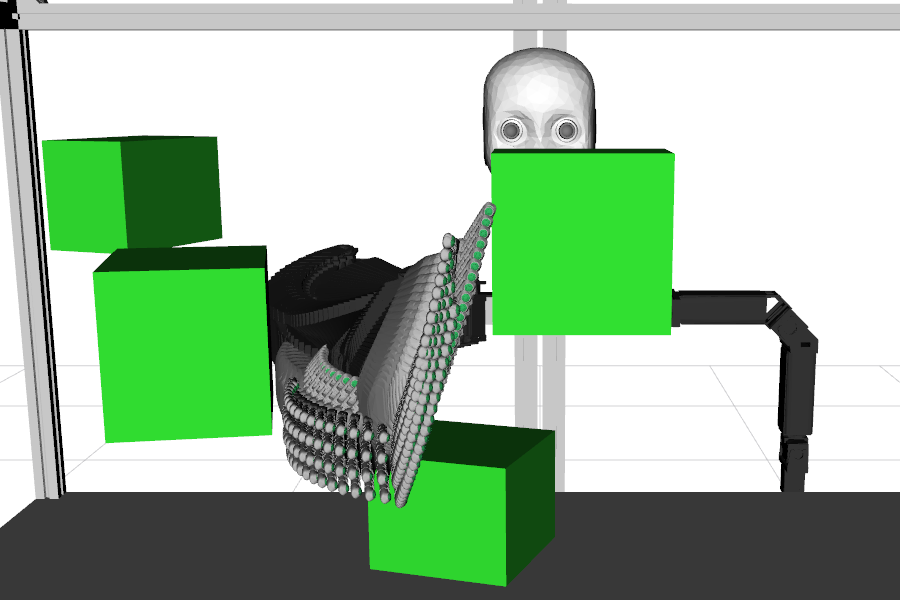}
\centering
\end{minipage}
\caption{Example keypoints and fixed step plans from the training dataset}
\label{fig:data_examples}
\end{figure}

In our experiments, we test two different plan representations shown in Fig~\ref{fig:data_examples}.
In the fixed step size representation, the robot configurations are spread along the path equally.
The keypoint representation reduces the plans to the key poses of the motion.
The key poses are retrieved by filtering the plans by an acceleration threshold, since a change in acceleration leads to a change in the direction of the motion.
The different representations of the plans strongly influence the distribution of the training data and, thereby, the model's behavior.

For each step along the plans, a training sample is created containing this step as the start, the last step of the plan as the goal, and the following 16 configurations as the target, which is padded with the goal.

\begin{table}[t]
\centering
\caption{The aggregated angular distance of the joints along the plans in radians.}
\begin{tabular}{ |c||c|c|c|c| } 
 \hline
 \textbf{Dataset} & \textbf{Mean}& \textbf{variance} & \textbf{max} & \textbf{min} \\ 
 \hline
 \hline
 \textbf{train} & 6.63 & 15.66 & 79.21 & 0.06  \\
 \hline
 \textbf{test} & 5.36 & 8.41 & 20.49 & 0.38  \\ \hline
\end{tabular}

\label{tab:dataset}
\vspace{-2em}

\end{table}

\begin{figure}[t]
    \centering
    \includegraphics[width=.8\textwidth]{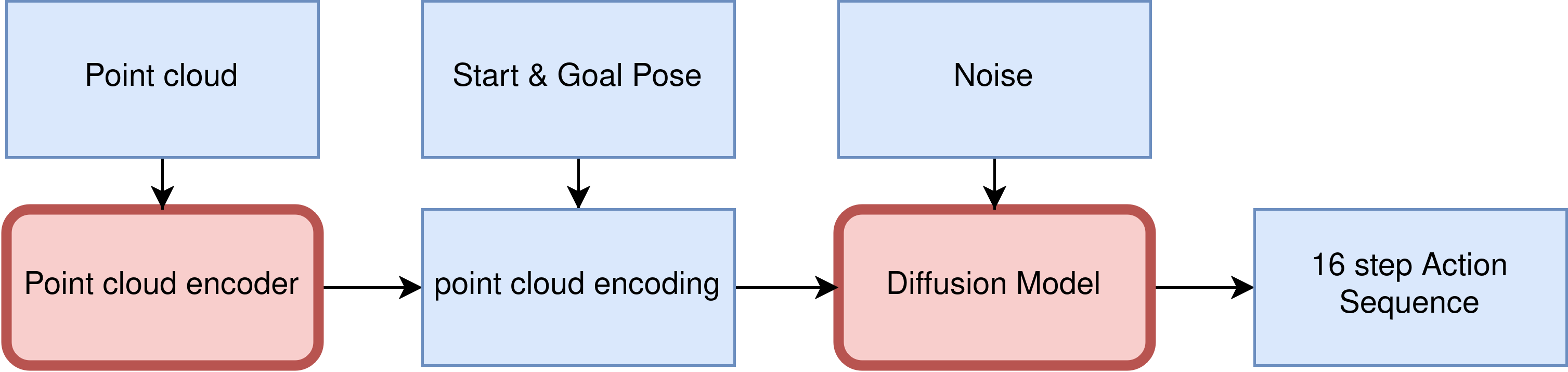}
    \caption{Neural Architecture: The point clouds of the environment are encoded and concatenated with the start and goal configuration. This vector is used as conditioning of the diffusion model, which generates a 16-step action sequence from a noise vector.}
    \label{fig:neural_arch}
\end{figure}

\subsection{Neural Architecture}
The model's architecture consists of a point cloud encoder and a diffusion-based action generator as shown in Fig.~\ref{fig:neural_arch}.
The initially hypothesized point cloud embedding is generated from the point cloud encoder and concatenated with the start and end states of the desired plan to create the condition of the diffusion model.

\subsubsection{Action Predictor}
The action prediction model is based on the diffusion policy implementation from \cite{chi2023diffusionpolicy}.
The CNN-based model uses a noise-predicting Unet that consists of multiple conditional residual blocks. 
The conditioning is achieved via Feature-wise Linear Modulation (FiLM) \cite{perez2017filmvisualreasoninggeneral}.
The action predictor generates 16-step action sequences for a given task description consisting of start, goal, and point cloud embedding.
To achieve this, the noise-prediction network is used as the gradient field of our training distribution.
The noise-prediction network takes a batch of action sequences of the form Bx16x8 as input and outputs a tensor of the same shape.
Since we treat the model as a gradient field, the output tensor is the gradient that moves our input vector to low-energy regions.
We can apply this gradient to the input to move it closer to the training distribution.
At inference time, we start the process from uniformly distributed noise and apply the inverted diffusion process 100 times to get the final output.

\begin{figure}[b]
    \centering
    \includegraphics[width=0.94\linewidth]{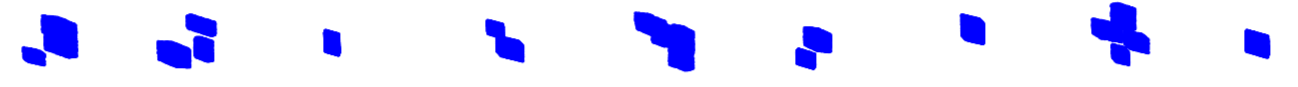}
    \includegraphics[width=0.955\linewidth]{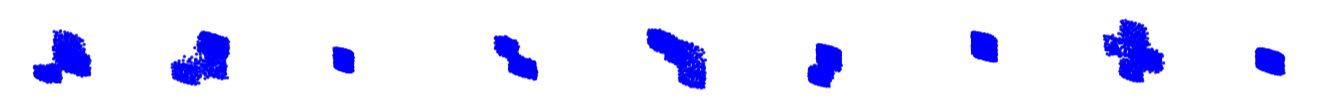}
    \caption{Example output of the auto encoder, input point clouds in the top row and model output in the bottom row.}
    \label{fig:pc_encodings}
\end{figure}

\subsubsection{Point cloud encoder}
The initially pursued, optional point cloud encoder is a simple PointNet \cite{qi2017pointnetdeeplearningpoint} based architecture using a stack of three 1D convolutional layers followed by a global max pooling operation to calculate the encodings.
The model takes fixed-size point clouds containing 4096 points.
The encoder is pre-trained in combination with an MLP decoder as an autoencoder on a task-specific dataset of point clouds.
The model is trained on the chamfer distance loss between the generated point cloud and the target point cloud.

Fig.~\ref{fig:pc_encodings} shows the testing output of the autoencoder model.
Although the recreated point clouds are noisier than the original ones, the model can capture the general geometric information of the point clouds.
We decided to use a simple architecture for our approach for multiple reasons.
Many modern architectures are designed to combine geometrical information with other features like point color, but since we only use geometrical data, we do not need this kind of capability.
Since computation time matters a lot in our approach, we decided to use a fast and time-tested approach for this work. 

\subsection{Trajectory generation}
Trajectories are generated by the sampling process of the diffusion model.
The global condition features are generated by concatenating the start and goal configuration in joint space and the point cloud embedding.
The inverse diffusion process is performed on uniformly sampled noise with the given condition to generate the next configurations along the trajectory.
The action horizon of the model is set to 16, which allows us to generate 16 steps of the trajectory for a given task description in one forward pass.
The generated plans are individually interpolated to a fixed step size, re-scaled, and checked for collision.
If the last configuration of the generated trajectory does not reach the target, we continue planning by using the final configuration of the generated plan as the new start configuration.

The neural approach offers batched inference, which can be used for batched trajectory generation.
This allows us to generate plans for different scenarios simultaneously or to create multiple plans for the same scenario.
In our approach, we are using the latter to increase and stabilize the performance of our model.
Since the diffusion process starts from noise, the model is stochastic, and sampling different noise for the same task can result in different outputs.

\begin{figure}[t]
   \centering
   \begin{minipage}{0.4\linewidth}
    \includegraphics[width=\linewidth]{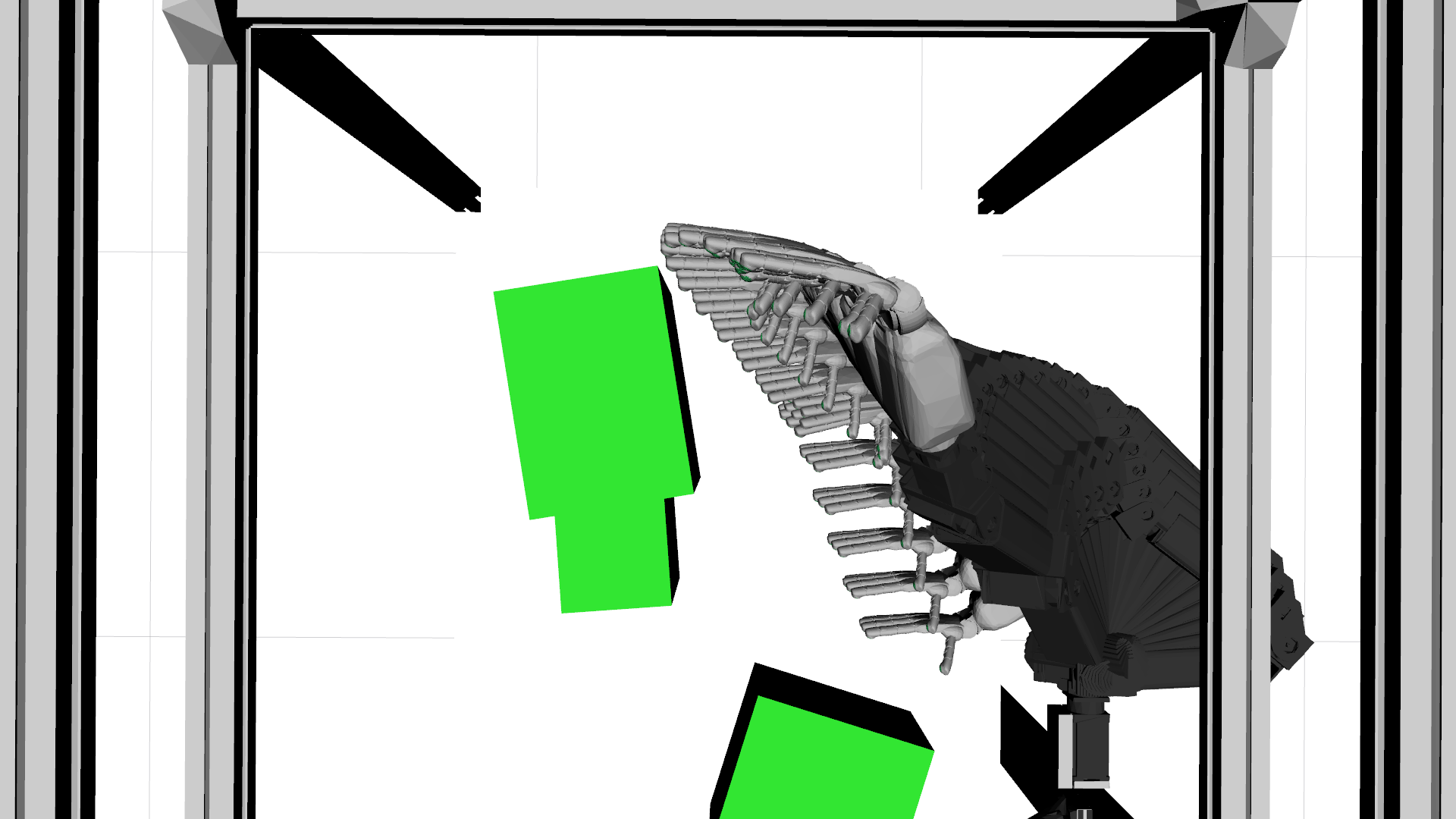}
    \centering
    \textbf{(a)}
    \end{minipage}
   \begin{minipage}{0.4\linewidth}
    \includegraphics[width=\linewidth]{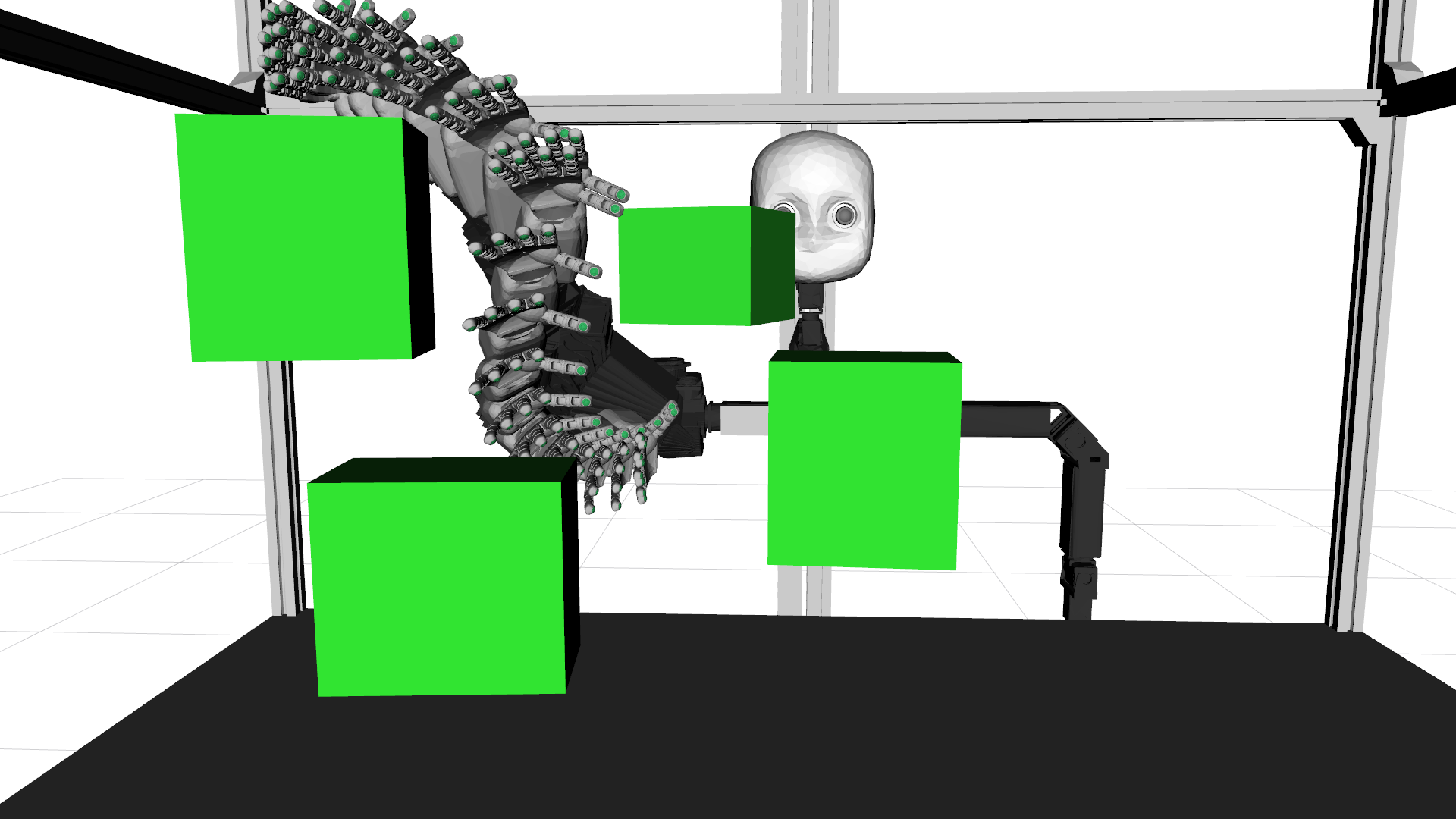}
    \centering
    \textbf{(b)}
    \end{minipage}
   \begin{minipage}{0.4\linewidth}
    \includegraphics[width=\linewidth]{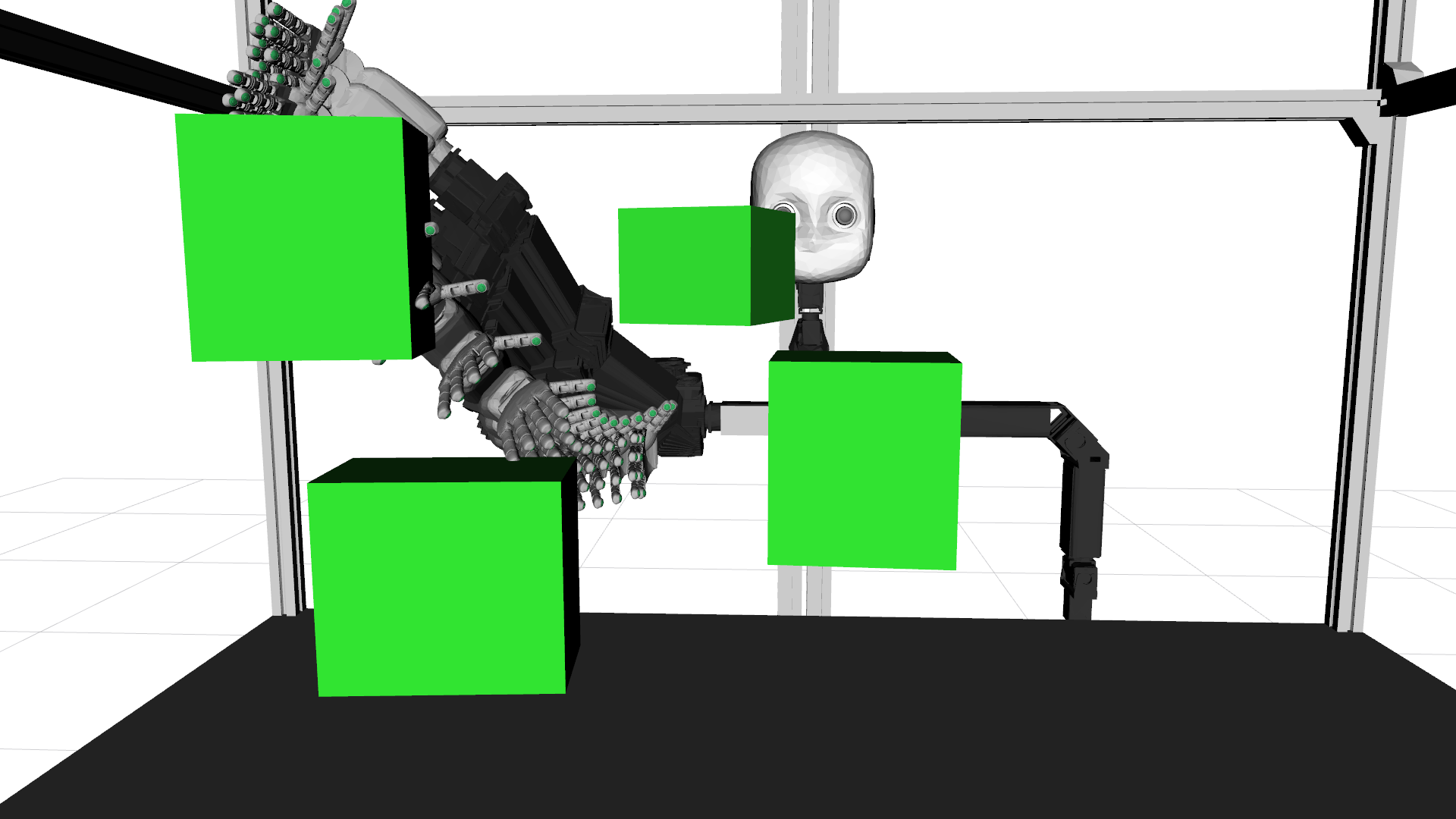}
    \centering
    \textbf{(c)}
    \end{minipage}
   \begin{minipage}{0.4\linewidth}
    \includegraphics[width=\linewidth]{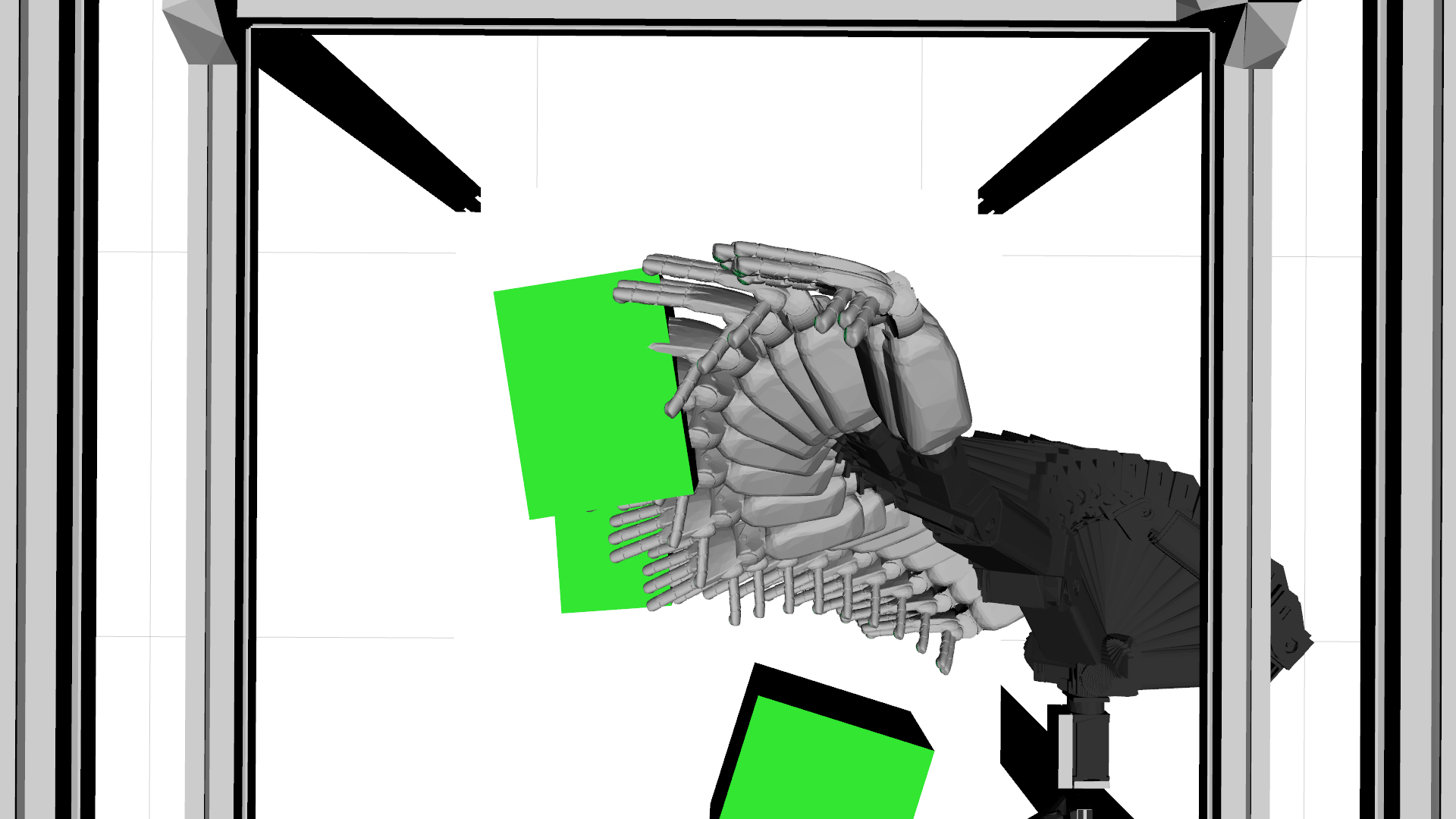}
    \centering
    \textbf{(d)}
    \end{minipage}
    \caption{Example of collision free solutions \textbf{(a)(b)} and colliding solutions \textbf{(c)(d)}}
    \label{fig:hard_example2}    
\end{figure}

\section{Results}
We trained the model on the keypoint plans and the fixed step representation plans and compared the model's behavior.
The batched planning approach is successful if at least one plan from the batch reaches the target with a collision-free trajectory.
Tab.~\ref{tab:performance} lists the success rates of our models on the test dataset.
Since our dataset contains many short plans, we report the success rate on the subset of hard trajectories with more than four keypoints separately.
We defined the model's success rate as the percentage of test plans the batched planning approach can solve successfully.
Fig.~\ref{fig:hard_example2} shows example plans generated by the model.
While the successful solutions create a nearly optimal motion around one of the obstacles, other plans from the same batch ignore the obstacle.
Since this hints at a low effect of the point cloud embeddings on the plan prediction, we conducted an ablation study on the model without the point cloud embeddings.
The high amount of short plans in the dataset introduces a strong bias toward direct motion towards the goal configuration in the training data.
To test the effect of this, we trained the model on the refined subset of the training plans with more than four keypoints.

\begin{table}[b]
\centering
\caption{Overview of model performance}
\begin{tabular}{ |c||c|c|c| } 
 \hline
 \textbf{Model} & \;\textbf{Success rate all}\; & \;\textbf{Success rate hard}\; \\ 
 \hline
 \hline
 \textbf{Fixed step} & 87.75\%  &  69.86\%\\
 \hline
 \textbf{Keypoints} & 90.50\% & 75.34\%  \\
 \hline
 \textbf{Ablation} & 88.00\% & 69.17\%  \\ 
 \hline
  \textbf{Keypoints refined} & 91.17\% & 77.40\%  \\ 
 \hline
\textbf{Ablation refined} & \textbf{92.00}\% & \textbf{78.67}\% \\ 
 \hline
\end{tabular}

\label{tab:performance}
\vspace{-2em}
\end{table}

\subsection{Inference Time}
The inference runtime of most diffusion models is higher compared to regression approaches that only require a single forward pass. However, the runtime of numerical methods is magnitudes higher due to extensive collision checking and iterative trajectory optimization.
During data generation, we fixed the planning time to 20 seconds to achieve an acceptable success rate of the numerical planner.
We report the average runtime for our motion planning architecture as well as the individual ratio of model inference runtime and collision checking runtime in Table~\ref{tab:Time}. The diffusion model has an average runtime of approximately 2 seconds, showing a very small variance, while the collision check has a runtime of 1 second.

\noindent
\begin{table}[t!]

\begin{minipage}{.4\textwidth}
\caption{Inference time per planning step in seconds (Keypoints)}    
    \centering
    \begin{tabular}{|c||c|c|c|c|}
        \hline
         \textbf{Time} & \textbf{mean} & \textbf{var} & \textbf{max} & \textbf{min} \\
        \hline
        \hline
        \textbf{Full Time} & 3.02 & 0.26 & 4.77 & 2.06\\
        \hline
         \textbf{Inference} & 1.97 & 0.01 & 2.41 & 1.86 \\
        \hline
         \textbf{Collision checking} & 1.05 & 0.22 & 2.83 & 0.09 \\
        \hline
    \end{tabular}
    
    \label{tab:Time}
\end{minipage}
\hspace{0.15\textwidth}
\begin{minipage}{.4\textwidth}
    \vspace{-11pt}
    \centering
    \caption{Difference in Plan length to test plans in radians}
    \begin{tabular}{|c||c|c|c|c|}
        \hline
         \textbf{Model} & \textbf{mean} & \textbf{var} & \textbf{max} & \textbf{min} \\
        \hline
        \hline
         \textbf{Fixed step }& 0.62 & 0.6 & 7.2 & -7.33 \\
        \hline
         \textbf{Keypoints} & 0.5 & 0.89 & 5.51 & -7.92 \\
        \hline
    \end{tabular}
    \label{tab:length_fixed_key}
\end{minipage}
\end{table}

\subsection{Fixed-step and Keypoint Representations}
\begin{figure}[t]
\begin{minipage}[c]{.49\linewidth}
\includegraphics[width=\linewidth]{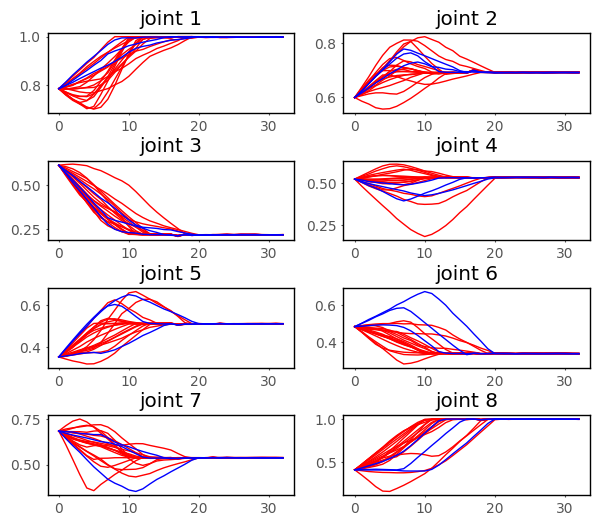}
\centering
\textbf{(a)}
\end{minipage}
\begin{minipage}[c]{.49\linewidth}
\includegraphics[width=\linewidth]{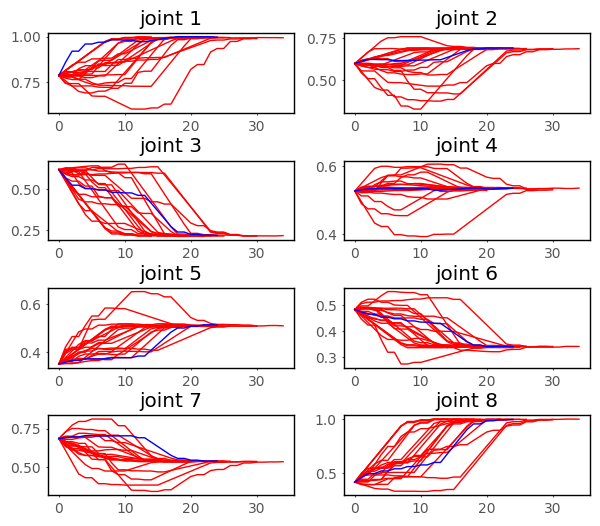}
\centering
\textbf{(b)}
\end{minipage}
\caption{Batch of plans in normalized joint space. Collision-free plans are marked blue. \textbf{(a)} Fixed step size plans \textbf{(b)} Keypoint plans}
\label{fig:jointspace}
\end{figure}

Fig.~\ref{fig:jointspace} shows example outputs of the two different approaches of plan representation.
The fixed-step approach generates smoother trajectories, imitating the smooth trajectories from the training data.
The key point trajectories are more jagged since we interpolate linearly between the key points.
The key point-based plans are less aligned since some of the plans include extra steps at the start.
Overall, the models show similar behavior and explore a similar part of the solution space.
As Tab.~\ref{tab:length_fixed_key} shows, fixed step plans tend to be slightly longer, which can be explained by the curve of the trajectories.
Since the fixed step size approach usually requires two planning steps to generate the full trajectory, which doubles the overall runtime, we focus further experiments on keypoint trajectories.

\subsection{Point cloud ablation}
\label{sec:ablation}
The point cloud is removed from the FiLM conditioning for this experiment. We aim to identify biases in the training data and process. We hypothesized that a lack of perception would lead to a decrease in performance since obstacle avoidance requires a representation of the search space. Surprisingly, the ablation model performed equally well compared to the standard model. 

The behaviour pattern originates from the batched path sampling process we apply to the stochastic diffusion model so that the model effectively learns to find a variety of paths for the same start and goal pose during training. It has to be noted that NICOL's large redundant configuration space is heavily constrained in this experiment, mainly due to the tabletop setup and the fixed hand orientation. This allows the training to collapse into a limited number of dynamic motion primitives accounting for a high success rate that does not generalize to the set of the most complex test scenes.

Fig.~\ref{fig:hist_sample}a and Fig.~\ref{fig:hist_sample}b show the model's performance with and without point clouds after 50 training epochs. Both models have a similar overall success rate of approximately 90\% on all plans, with the point cloud-conditioned model showing a better performance on the subset of more difficult plans Tab.~\ref{tab:performance}. When we analyze the distribution of likelihoods to generate a successful solution after 50 epochs, both distributions look very similar, with the model based on the point cloud encodings having a higher in-batch success rate on the easy trajectories. While the point cloud embeddings did not improve the success rate, they affected the plan length positively, as Tab.~\ref{tab:nopc_length} shows.

\begin{table}[t]
\begin{minipage}{.5\linewidth}
    \centering
    \caption{Difference in plan length to \\ test plans in radians}
    \begin{tabular}{|c||c|c|c|c|}
        \hline
         \textbf{Model} & \textbf{mean} & \textbf{var} & \textbf{max} & \textbf{min} \\
        \hline
        \hline 
       \textbf{Keypoint} & 0.5 & 0.89 & 5.51 & -7.92 \\
        \hline
       \textbf{Ablation}& 0.66 & 0.99 & 7.94 & -5.42 \\
        \hline
    \end{tabular}
    
    \label{tab:nopc_length}
\end{minipage}
\begin{minipage}{.5\linewidth}
    \centering
     \caption{Refined dataset: Difference in plan length to test plans in radians}
    \begin{tabular}{|c||c|c|c|c|}
        \hline
         \textbf{Model} & \textbf{mean} & \textbf{var} & \textbf{max} & \textbf{min} \\
        \hline
        \hline 
        \textbf{Keypoint} & 0.74 & 0.72 & 3.99 & -7.07 \\
        \hline
        \textbf{Ablation} & 1.01 & 1.22 & 9.04 & -7.02 \\
        \hline
    \end{tabular}
   
    \label{tab:my_label}
\end{minipage}
\end{table}

\subsection{Refined Training data}
As our experiments have shown, the model learns to solve easy plans with high probability, while struggling with the harder tasks from the test set.
Since our dataset is biased towards short trajectories, removing these from the training data should help the model generalize better towards the harder tasks.
We can filter the training data for all the plans that move directly towards the target, which we consider easy, and remove them from the training dataset.
We define these plans to have four or fewer key points.

\begin{figure}
\begin{minipage}[c]{.24\linewidth}
\centering
\label{fig:}
\includegraphics[width=\linewidth]{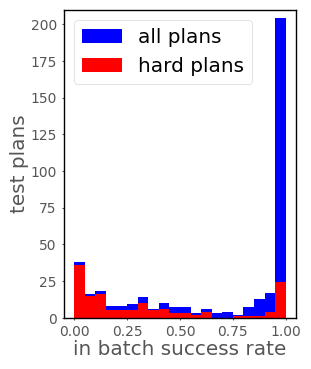}
\textbf{(a)}
\end{minipage}
\begin{minipage}[c]{.24\linewidth}
\centering
\includegraphics[width=\linewidth]{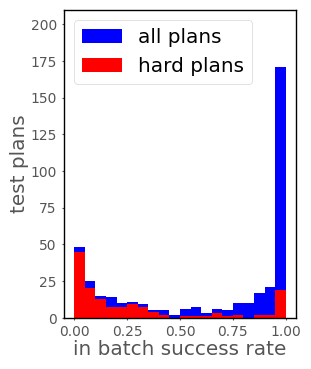}
\textbf{(b)}
\end{minipage}
\begin{minipage}[c]{.24\linewidth}
\centering
\includegraphics[width=\linewidth]{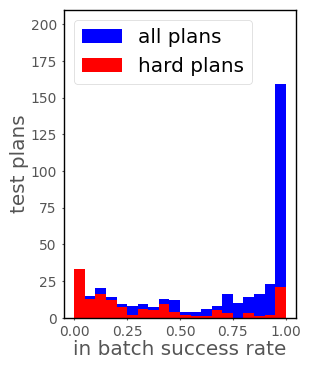}
\textbf{(c)}
\end{minipage}
\begin{minipage}[c]{.24\linewidth}
\centering
\includegraphics[width=\linewidth]{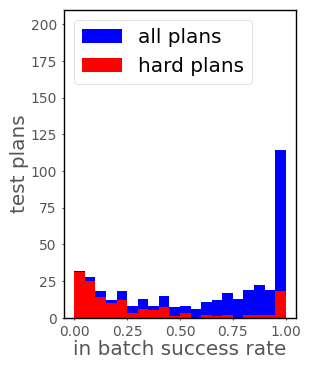}
\textbf{(d)}
\end{minipage}
\caption{In batch success rate \\ \textbf{(a)} baseline \textbf{(b)} ablation \textbf{(c)} baseline refined dataset \textbf{(d)} ablation refined dataset}
\label{fig:hist_sample}
\end{figure}

Fig.~\ref{fig:hist_sample}c and \ref{fig:hist_sample}d show the behavior of the models with and without point clouds, trained on the refined dataset. 
Reducing the bias in the training data positively affected the model's performance.
A distribution shift toward the middle can be observed by looking at the distribution of likelihoods.
While some certainty for the easy plans was lost, the model was able to find collision-free solutions for more of the hard scenarios.
It can be seen in Tab.~\ref{tab:performance} that the model generalizes to the easy trajectories, even though not being trained on them.

When we examine the plan length metrics, we see that all models lost a bit of average plan quality by removing the easy samples from the training data.
This lines up with our observations of the distribution of the in-batch success being less concentrated at the extremes.
 
\section{Discussion}
The behavior cloning approach we use does not include collision information in the training loss. The network is merely trained on successful trajectories that are derived from a stochastic algorithm and is thus biased by many factors, such as the planning algorithm, the collision checking algorithm, and the joint limits of the robot. A suitable, differentiable, batched collision checker, which is, to the best of our knowledge, unavailable at this point in time, would pose a new gold standard to enforce point cloud-conditioning in a reinforcement learning-inspired manner.
Related work \cite{fishman2022motionpolicynetworks,dalal2024neuralmpgeneralistneural} shows that the integration of point clouds into action models is 1) successfully possible and 2) crucial for successful deployment in real-world scenarios. The cited approaches generate more realistic scenes, such as household environments, to derive task-specific plans, moving away from the definition of generalized motion planning. In addition, these approaches use larger datasets by an order of magnitude. 
We identify the fixed end-effector orientation in our data aggregation method as a key issue that can lead to a missing variety of the dataset. Motion Policy Networks \cite{fishman2022motionpolicynetworks} use point cloud representations of the robot and the end-effector target in the same latent space as the scene point clouds in combination with the robot state and report a loss of 30\% success rate when using only the joint state instead. We identify the robot proprioception via point cloud encodings as the key advantage compared to our approach, which only uses point cloud encodings of the environment.

In the presented approach, we successfully introduce a neural model that generates collision-free trajectories much faster than traditional planners.
The results of our point cloud ablation study revealed that the point cloud embeddings in the input, while slightly reducing the average plan length, do not increase the success rate of our model, even though the point cloud auto-encoder we utilize produces meaningful embeddings.
The architecture of our ablation study is similar to the approach of Carvalho et al. \cite{carvalho2023motion}, who observed similar capabilities of diffusion models for motion planning.
The simpler architecture does not directly rely on the perception of the robot, bringing the model to a real robot is therefore much easier, since we do not have to accommodate for the shift in the input distribution usually introduced by noisy sensors.
The average runtime of our model of 3 seconds is by an order of magnitude lower compared to the 20-second planning time of the PRM* planner, while reaching a success rate of 90\% on the test set.

\section{Conclusion}
In this paper, we introduced a novel diffusion architecture for robotic motion planning.
While our approach shows good performance on the test data, the integration of point clouds into the sampling process was still challenging in our experiments, as shown in the ablation study.
We suspect various biases in the dataset and the missing proprioception of the robotic agent as the origin of the malfunctioning of the point cloud input, which we are going to investigate further in our future work.
By facilitating the batched plan generation capabilities of our model and the use of keypoint-based plans, our approach is an order of magnitude faster than the planning process of numerical solvers as PRM*, while keeping a success rate of 90\% on the test data.
Combining our approach with a numeric planner as a backup allows us to skip the lengthy planning most of the time at a low constant cost while keeping a high quality of plans in terms of plan length. 

%
%
%
\bibliographystyle{splncs04}
\bibliography{bib}
\end{document}